\documentclass[12pt]{article}

\usepackage{preamble}

\title{\Large \textbf{The Berkeley Single Cell Computational Microscopy (BSCCM) Dataset}}

\date{} 

\author[1,2,3]{Henry Pinkard}
\author[4]{Cherry Liu}
\author[4]{Fanice Nyatigo}
\author[4]{Daniel A. Fletcher}
\author[1,2]{Laura Waller}

\affil[1]{Department of Electrical Engineering and Computer Sciences, University of California, Berkeley}
\affil[2]{Berkeley Institute for Data Science}
\affil[3]{University of California, San Francisco Bakar Computational Health Sciences Institute}
\affil[4]{Department of Bioengineering, University of California, Berkeley}

\begin{document}

\maketitle

\renewcommand{\abstractname}{\vspace{-26pt }} 

\begin{abstract} \bf

Computational microscopy, in which hardware and algorithms of an imaging system are jointly designed, shows promise for making imaging systems that cost less, perform more robustly, and collect new types of information. Often, the performance of computational imaging systems, especially those that incorporate machine learning, is sample-dependent. Thus, standardized datasets are an essential tool for comparing the performance of different approaches. Here, we introduce the Berkeley Single Cell Computational Microscopy (BSCCM) dataset, which contains over ~12,000,000 images of 400,000  of individual white blood cells. The dataset contains images captured with multiple illumination patterns on an LED array microscope and fluorescent measurements of the abundance of surface proteins that mark different cell types. We hope this dataset will provide a valuable resource for the development and testing of new algorithms in computational microscopy and computer vision with practical biomedical applications.

\end {abstract}

\section{Introduction}

As the number and variety of computational microscopy techniques continues to grow, standardized performance benchmarks are essential for both guiding new research and understanding the utility of new techniques. This need has been compounded by the fact that, like many other fields that rely on image processing, the past decade has seen an explosion of techniques that use data-driven machine learning methods, the most prominent example being deep neural networks \cite{Lecun2015} . 

Since data-driven machine learning models depend heavily on the data fed into them, developing and benchmarking new algorithms depends upon the existence of easily-accessible standardized datasets. The benefits of such datasets are two-fold: they allow new advances to be compared against existing approaches without the confounding effects of different data, and they speed the development of new techniques by freeing researchers from the burden of collecting and processing their own data. In the computer vision field, datasets such as MNIST~\cite{LeCun1998} and ImageNet~\cite{Deng2009} provide simplified, easily understood problems (handwritten digit and natural image classification, respectively), and algorithms developed using these datasets have gone on to be quite successful at generalizing to a variety of diverse applications from diagnosing skin cancer~\cite{Esteva2017} to mapping poverty from satellite imagery~\cite{Babenko2017}.

Unlike most computer vision techniques, computational microscopy encompasses the design of imaging systems, in addition to the processing of the images they produce. This often requires formulating models of the imaging system's physics and/or acquiring additional calibration data. For example, deconvolution algorithms usually require knowledge of the system's point spread function (PSF); Diffraction tomography, in which a 3D image of a sample is reconstructed from 2D projections, requires a physical model of how those projections were formed. It is in some cases possible to learn these extra parameters by estimating them alongside the image reconstruction (i.e. blind deconvolution~\cite{Levin2011} or self-calibration for illumination angles~\cite{Eckert2018}). However, the development of such self-calibration algorithms would similarly benefit from reference datasets with available ground truth.

Often, the benchmarks used in the creation of new computational microscopy algorithms bear little similarity to their possible applications. For example, a common strategy is to measure algorithm quality by presenting a an image of the USAF resolution target. While useful as a starting point, the potential to readily evaluate new algorithms on scenarios closer to real-world applications is a desirable feature, considering the sample-dependent performance of computational imaging systems. However, the difficulty of collecting and processing such datasets often precludes this in practice.

To address this need, we created the Berkeley Single Cell Computational Microscopy (BSCCM) dataset, which:

\begin{itemize}
    \item is large enough to be used as training data on contemporary deep neural networks 
    \item contains the structured metadata and calibration required by computational microscopy algorithms 
    \item is directly relevant to one of the most commonly performed laboratory tests: counting different types of white blood cells
\end{itemize}

The dataset contains $400,000$ images of single white blood cells taken with an LED-array microscope \cite{Zheng2011, Phillips2016}, as well as measurements of the same cells with the two traditional measurements used in phenotyping assays: histology staining and measurement of marker proteins via fluorophore-conjugated antibodies. The fluorescent images of antibody-stained cells were further processed to provide expression levels of different proteins on individual cells, and provide cell-type labels for different cells. The latter provides a way to use the dataset for cell type classification problems. 

\begin{figure*}[htb]
\centering
\includegraphics[width=0.8\linewidth]{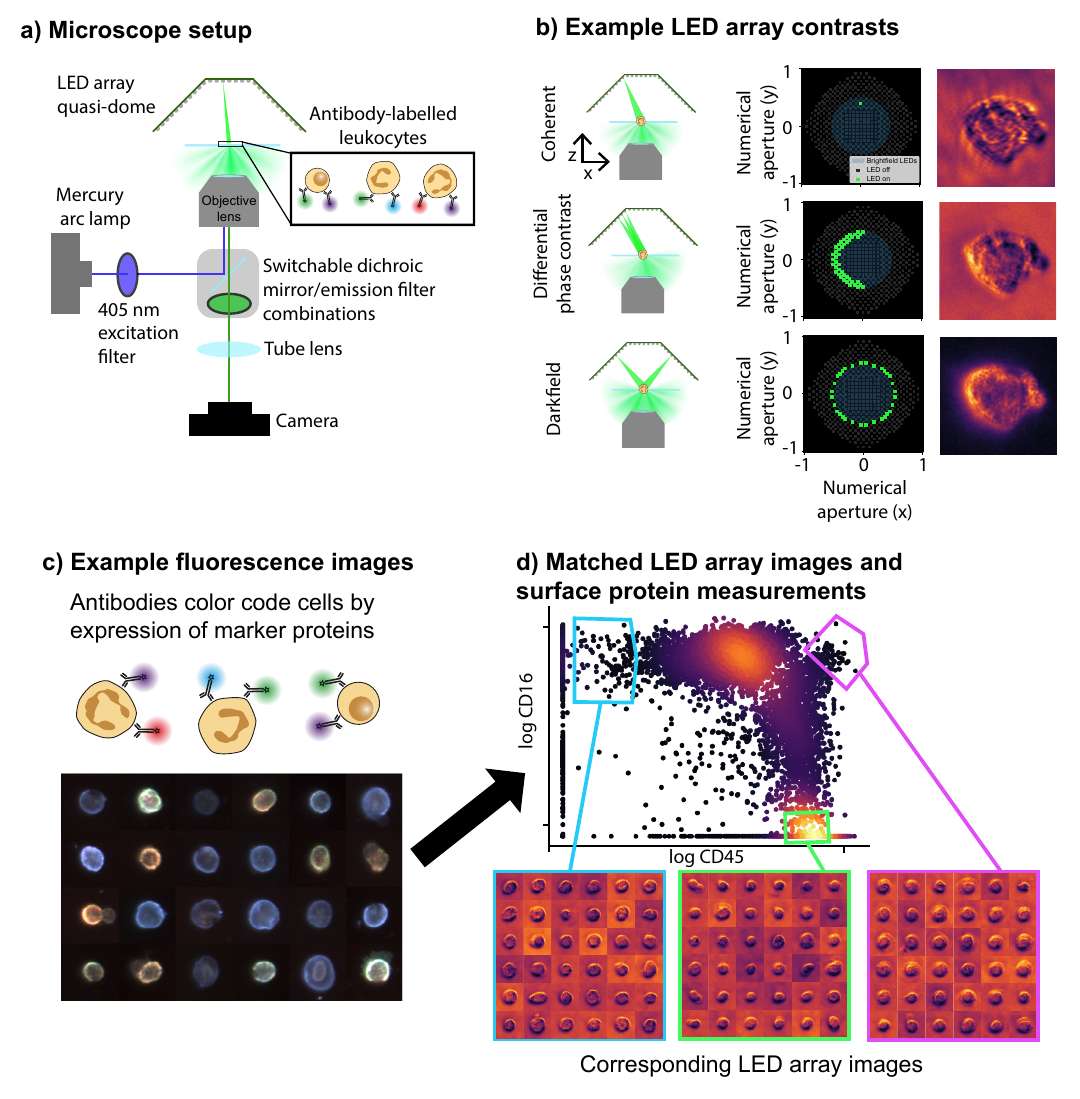}
\caption{\textbf{Berkeley Single Cell Computational Microscopy (BSCCM) dataset overview.} \textbf{a)} Schematic of the microscope used in data collection: a commercial body fluorescence microscope with its trans-illumination lamp replaced with a programmable LED array quasi-dome. \textbf{b)} The LED array was used for label-free imaging of cells with different illumination patterns. \textbf{c)} The fluorescence light path was used capture 6-channel fluorescence images on the same cells. \textbf{d)} This provided both protein expression levels and label-free images on the same cells.}
\label{overview}
\end{figure*}


\section{Background}
\label{section_background}

\subsection{LED array microscopy}

Label-free microscopy generates contrast from intrinsic properties of samples instead of exogenous chemical contrast agents (such as light absorbing or emitting dyes that bind to specific molecules). Although label-free techniques have been around for close to a century~\cite{Zernike1955}, recently advances have enabled the extraction of quantitative estimates of physical properties, such as the phase delay caused by the sample~\cite{Popescu2015} and polarization~\cite{Oldenbourg2005}. As a result, physical properties like the relative protein/lipid content or the polymeric structure of molecules can be spatially resolved~\cite{Yeh_upti_2021, Mir2011a}, providing useful phenotypic information about biological samples. 

The LED array microscope~\cite{Zheng2011} offers a simple yet powerful system for label-free imaging of biological samples. It is realized by replacing the illumination lamp on a traditional, transmitted light microscope with a programmable LED array, so that the sample can be illuminated with different angles of light by turning on different LEDs. By illuminating with different patterns, it is possible to utilize several different contrast generation mechanisms, including brightfield, darkfield, and differential phase contrast, on a single system without any moving parts.

This can done using a planar LED array or, as in our case, an LED array quasi-dome~\cite{Phillips2016}. The advantage of the latter is that it yields increased intensity at high illumination angles compared to the planar array. Specifically, the intensity of incident light on the sample given an angle $\theta$ relative to the optical axis is proportional to $\frac{1}{cos^4\theta}$ for the planar array, whereas the quasi-dome's dropoff with angle is $\frac{1}{cos\theta}$.

By capturing images with different illumination patterns and using physics-based models of the image formation process, a variety of techniques can be implemented. These include multi-contrast imaging \cite{Zheng2011, Liu2014d}, the synthesis of high-resolution images from low-resolution inputs via Fourier Ptychography \cite{Zheng2013, Ou2013, Tian2015e}, estimation of 3D structure from 2D measurements \cite{Tian2014DPC3D, Tian2015b, Horstmeyer2015, Ling2018}, quantitative phase imaging \cite{Ou2013, Tian2015c}, and digital aberration correction \cite{MChen2018, Eckert2018}. 

Achieving optimal performance with these techniques usually requires the use of pre-processing steps such as shading corrections, in which raw measurements are divided by an image taken on the same system under the same illumination, but without a sample. This pre-processing accounts for inhomogenous illumination caused by imperfections in the optical system. As a result, it is important for standardized datasets to include the calibration images needed to make such corrections.

The physics of LED array illumination provides a convenient way to test different illumination strategies \textit{in silico}: Since each LED is incoherent with all other LEDs, images corresponding to illumination with multiple patterns can be synthesized \textit{in silico} by adding together the corresponding single-LED illuminated images. Though these two situations are not perfectly equivalent due to factors such as the camera offset, read noise, etc., they are practically similar enough that computationally-optimized illumination in fact does generalize to improved imaging on experimental systems \cite{Horstmeyer2017a, Kellman2019}.

\subsection{Why single white blood cells?}

A desirable feature of a standardized benchmark dataset is its relevance to real-world applications. Among potential biomedical applications, imaging single cells is particularly appealing for three reasons. First, single-cell phenotyping assays already have ubiquitous clinical application. For example, counting the number of different white blood cell types in a blood sample from a patient (peripheral blood leukocyte differential count) is one of the most frequently used clinical laboratory tests~\cite{Buttarello2008}. Furthermore, cutting edge methods that measure the RNA~\cite{Kharchenko2021} or protein~\cite{Kelly2020} composition of single cells continue to yield new biological knowledge and clinical applications. Second, label-free computational microscopy has the potential to augment or replace many single-cells methods (microscopy-based or otherwise), thanks to its speed, relatively low cost, and non-toxicity to cells. Third, many imaging techniques that can be performed using LED array illumination have already been shown to capture rich, biologically significant information for characterizing cells~\cite{Wang2011, Wilson2013, Chen2016, Eulenberg2016, Yoon2017}.

Realizing the full potential of label-free computational microscopy for single cell imaging relies upon finding ways to benchmark and then optimize the performance of these methods for biological assays. The majority of single cell assays capture information by measuring the levels of various known molecular markers on single cells (e.g. specific messenger RNAs or proteins). As a result, being able to compare the information captured by a label-free method with the information present in existing assays is an important step.

White blood cells (leukocytes) are an ideal model system for this task. They are clinically significant in a variety of disease processes \cite{Buttarello2008, Hensley2012}: changes in the frequencies of these cell types in peripheral blood can be informative about a number of diseases including infections, inflammatory disease, and cancer. They are morphologically-diverse, containing substantial variation in their size, the shape of their nuclei, and their cytoplasmic composition. Finally, there are multiple, ubiquitously-used assays for phenotypically characterizing them~\cite{Guy2017}: characterization of expression of proteins that serve as markers of cell lineage (usually based on antibody binding on a flow cytometer~\cite{Faucher2007, Autissier2010, Guy2017}) or using a histology stain and manual examination on a brightfield microscope. This redundancy provides multiple means of benchmarking the performance of label-free microscopy.

\section{Dataset overview}
\label{section_dataset_overview}
This section gives an overview of the different versions of the BSCCM dataset, with the aim of providing the most important information for getting started. 

All data were collected on a Zeiss Axio Observer microscope with its illumination source replaced by a programmable quasi-dome LED array~\cite{Phillips2016} (\textbf{Fig. \ref{overview}a}). This microscope was also equipped with an epifluorescence light path consisting of a mercury arc coupled to a 405nm bandpass filter, and 6 switchable dichroic mirror-emission filter combinations. This setup enabled label-free imaging of cells using the LED array (\textbf{Fig. \ref{overview}b}) and fluorescence imaging on the same cells (\textbf{Fig. \ref{overview}c}), which, after processing (\textbf{Section \ref{data_processing}}) enabled label free image to be correlated with expression levels of proteins marking cell type on the same cells.

After LED array/fluorescence imaging, a subset of cells was stained with a light absorbing histology stain (Wright's stain) and imaged again using the LED array using red, green, and blue brightfield patterns to produce an RGB image. The LED array and fluorescence images were collected using a $20\times$ 0.5 NA objective lens. The histology images were collected with a $63\times$ 1.4 NA oil immersion lens.

\begin{figure*}[htbp]
\centering
\includegraphics[width=\linewidth]{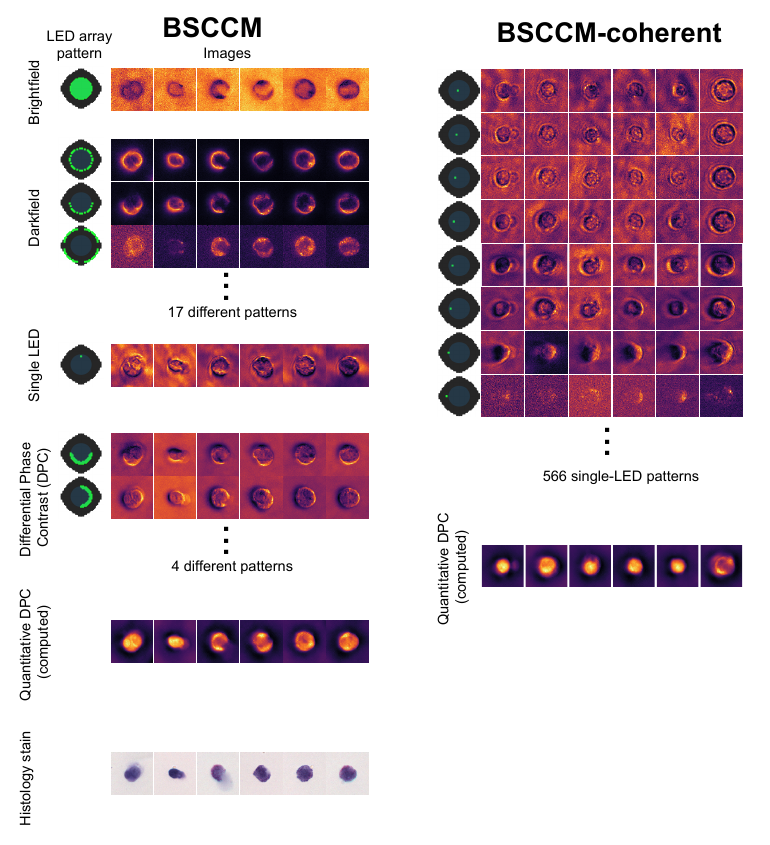}
\caption{\textbf{LED array and histology contrasts in BSCCM and BSCCM-coherent} }
\label{contrast_types}
\end{figure*}

\begin{figure*}[htbp]
\centering
\includegraphics[width=\linewidth]{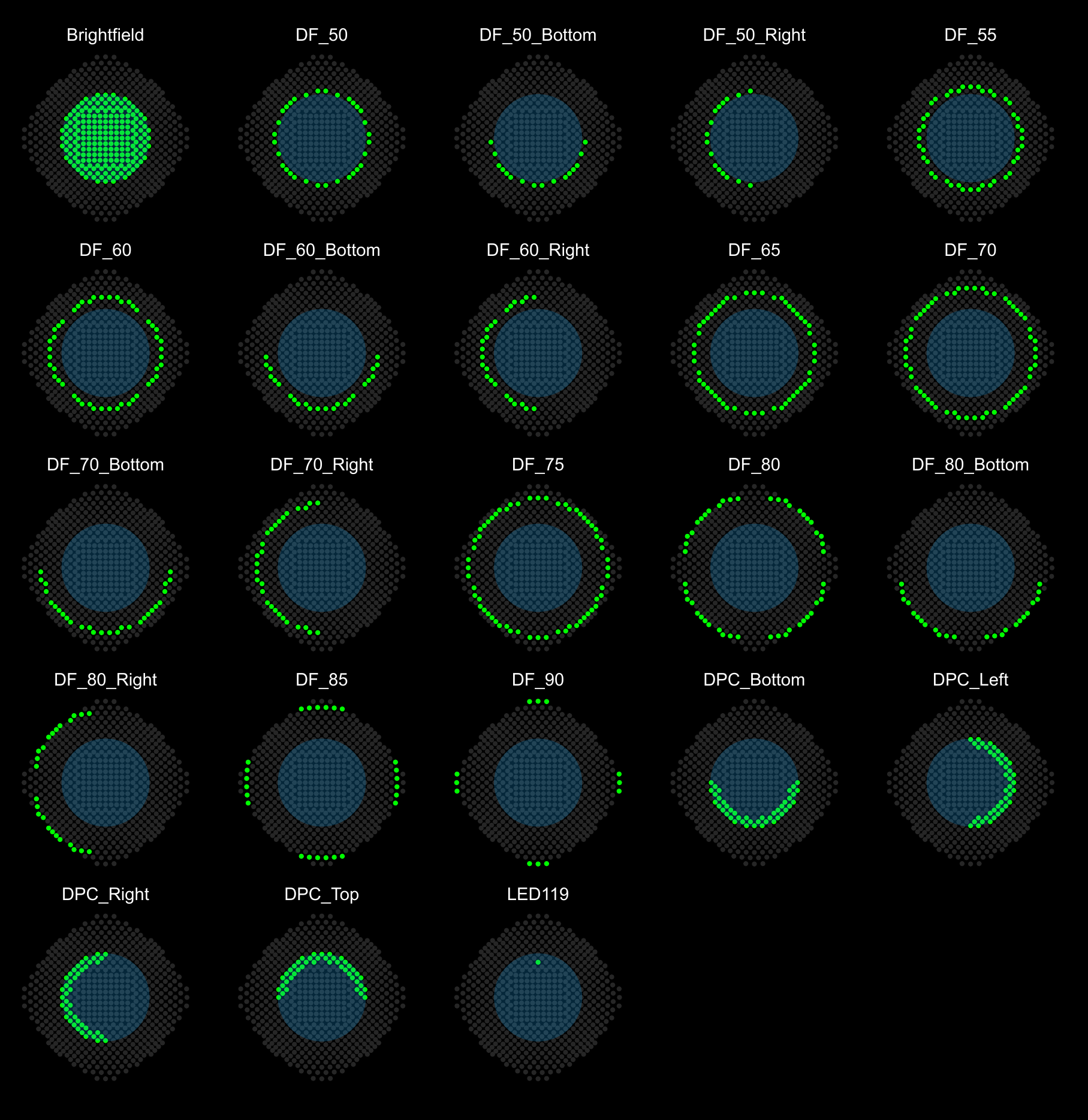}
\caption{\textbf{LED array patterns used in the BSCCM dataset and the names used for them in the dataset}}
\label{bsccm_led_patterns}
\end{figure*}

The BSCCM dataset comes in several different versions with different sizes (\textbf{Table \ref{dataset_size_comp}}) and use cases. The default version of the dataset (BSCCM) has 22 images of each cell corresponding to a different multi-LED pattern (brightfield, darkfield, or differential phase contrast) and a single image for each cell taken under illumination with a single LED (\textbf{Fig. \ref{contrast_types}}, \textbf{Fig. \ref{bsccm_led_patterns}}).

\begin{table}[h!]
    \centering
    \caption{\textbf{Comparison of BSCCM dataset image and data sizes}}
    \label{dataset_size_comp}
    \begin{tabular}{|l|r|r|c|}
    \hline
    \textbf{Name} & \textbf{\makecell{Size\\(GB)}} & \textbf{\# Cells} & \textbf{\makecell{Image Specifications\\\textcolor{blue}{LED-array/Fluor}\\ \textcolor{magenta}{Histology}}}\\
    \hline
    \textbf{BSCCM} & 228 & 412,941 & \makecell{\textcolor{blue}{128$\times$128} \textcolor{blue}{12-bit} \\ \textcolor{magenta}{398$\times$398} \textcolor{magenta}{36-bit}}\\ 
    \hline
    \textbf{BSCC\textcolor{OliveGreen}{MNIST}} & 58 & 412,941  & \makecell{\textcolor{blue}{28$\times$28} \textcolor{blue}{8-bit} \\ \textcolor{magenta}{28$\times$28} \textcolor{magenta}{24-bit}} \\
    \hline
    \textbf{BSCCM-coherent} & 30 & 4,304 & \makecell{\textcolor{blue}{128$\times$128} \textcolor{blue}{12-bit}} \\
    \hline
    \textbf{BSCCM-tiny} & 0.6 & 1,000  & \makecell{\textcolor{blue}{128$\times$128} \textcolor{blue}{12-bit} \\ \textcolor{magenta}{398$\times$398} \textcolor{magenta}{36-bit}}\\ 
    \hline
    \textbf{BSCC\textcolor{OliveGreen}{MNIST}-tiny} & 0.2 & 1,000  & \makecell{\textcolor{blue}{28$\times$28} \textcolor{blue}{8-bit} \\ \textcolor{magenta}{28$\times$28} \textcolor{magenta}{24-bit}} \\
    \hline
    \textbf{BSCCM-coherent-tiny} & 0.6  & 100 &  \makecell{\textcolor{blue}{128$\times$128} \textcolor{blue}{12-bit}} \\ 
    \hline
    \end{tabular}
\end{table}

\paragraph{BSCCM-coherent.}
Techniques such as diffraction tomography and Fourier ptychography often rely on measurements taken with spatially coherent plane wave illumination from different angles. This can be performed on the LED array by lighting up only a single LED at time. We collected a version of BSCCM specifically for this purpose, called BSCCM-coherent, which contains 566 different illumination patterns, each corresponding to illumination by a different single LED (\textbf{Fig. \ref{contrast_types}}, \textbf{Fig. \ref{bsccm_led_patterns}}).

A summary of the difference between the datasets in terms of the different contrasts they provide and their antibody staining conditions is given in \textbf{Table \ref{regular_vs_coherent_table}}.

\begin{table}[h!]
    \centering
    \caption{\textbf{Comparison of BSCCM dataset image types}}
     \label{regular_vs_coherent_table}
    \begin{tabular}{|l|r|r|c|c|}
    \hline
    \textbf{Name} & \textbf{LED array patterns} & \textbf{\makecell{Antibody \\stain conditions}} & \textbf{Histology stain} \\
    \hline
    \textbf{BSCCM} & \makecell{22 multi-LED + \\ 1 single-LED}  &
    \makecell{None \\ Single antibody \\ All antibodies}
    & Some cells  \\ 
    \hline
    \textbf{BSCCM-coherent} & \makecell{566 single-LED}  & 
    \makecell{None \\  All antibodies}
    & No \\
    \hline
    \end{tabular}
\end{table}

\paragraph{BSCC\color{OliveGreen}{MNIST}.}
Often when first setting up a machine learning pipeline, it is easiest to get started with a dataset of small images that can easily fit in memory of the training hardware (e.g. a GPU). MNIST~\cite{LeCun1998}, a dataset of 28x28 pixel images of handwritten digits is a good starting point. Given the large number of existing code examples built for 28x28 pixel images, we created versions of BSCCM with all images downsampled to 28x28. The BSCCMNIST datasets all contain the same contrast modalities and antibody staining conditions as their corresponding full versions.

\paragraph{BSCCM*-tiny.}
Each of BSCCM, $\text{BSCC\color{OliveGreen}{MNIST}}$, and BSCCM-coherent, has a corresponding "tiny" version, which is a subset of cells present in the full dataset. The purpose of these datasets is to provide an easy way to get started with the datasets on a laptop.

\subsection{Fluorescence}

\begin{figure*}[htbp]
\centering
\includegraphics[width=\linewidth]{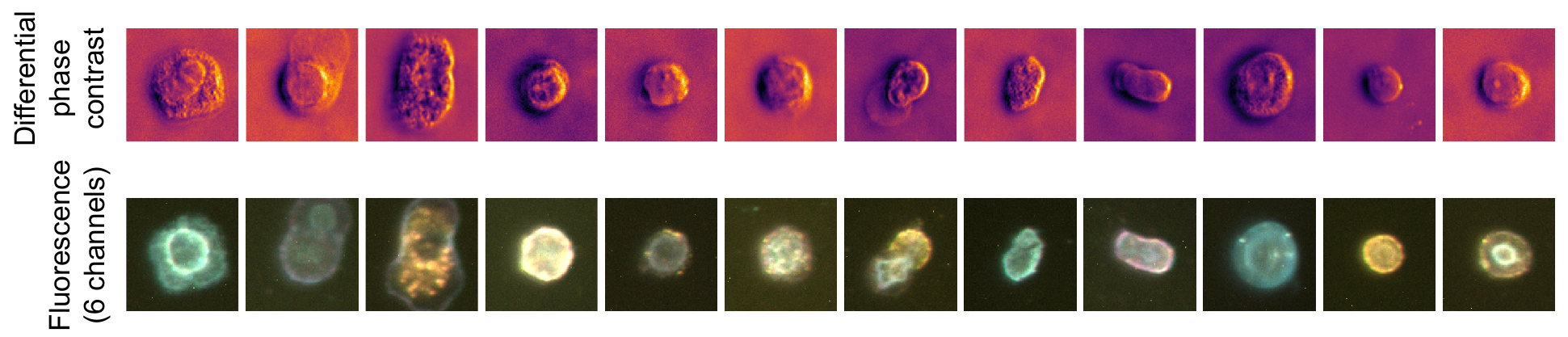}
\caption{\textbf{DPC and 6 channel fluorescence images in the all antibodies staining condition}}
\label{fluor_and_dpc_images}
\end{figure*}

\begin{figure*}[htbp]
\centering
\includegraphics[width=\linewidth]{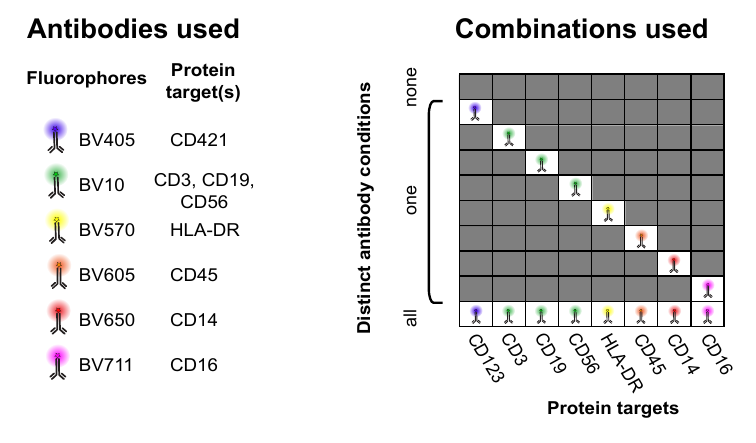}
\caption{\textbf{Antibody staining conditons}}
\label{antibody_conditions}
\end{figure*}

In addition to LED array illumination patterns, each dataset also contains fluorescence images in 6 channels (\textbf{Fig. \ref{fluor_and_dpc_images}}), though they differ in which antibodies were used (\textbf{Fig. \ref{antibody_conditions}}). For BSCCM and $\text{BSCC\color{OliveGreen}{MNIST}}$ these conditions include either a single antibody at a time, all antibodies together, or no antibodies. BSCCM-coherent contains only all-antibodies and no antibody conditions.

Since the fluorophore spectra overlap and bleed through into multiple channels, the raw fluorescence data was demixed (\textbf{Section \ref{demixing}}) to estimate the relative abundance of each antibody's protein target, the measurement of which provides a metric with visible morphological correlates (\textbf{Fig. \ref{overview}d})

\subsubsection{Cell type labels}
For a subset of cells that were stained with all antibodies, the protein abundance estimates were used to create cell type labels based on the cell's pattern of protein expression (\textbf{Fig. \ref{classification}}). Two versions of the labels are available: one that has 3 classes--lymphocyte, monocyte, and granulocyte)--and one that has 10--multiple subtypes of the previous 3 + red blood cells and 2 whose cell type is unclear. 

\subsection{Metadata and calibration}
In addition to the image data, the BSCCM datasets also contain detailed metadata corresponding to each cell. This includes information about the microscope: pixel sizes of the sensor, all wavelengths, angles of each LED, etc. There are also background images of each LED illumination pattern, computed by taking the pixel-wise median (smaller percentiles are available as well) across many fields-of-view, which effectively removes the contribution of the cells themselves and captures the microscope's illumination pattern.

\section{Accessing the data}
The raw data is available at \url{https://doi.org/10.5061/dryad.sxksn038s}. It can be downloaded and used with the \texttt{bsccm} python package, which is hosted at: \url{https://github.com/Waller-Lab/BSCCM}

\section{Acknowledgements}
The authors thank Brian Belardi for many helpful discussions and feedback in designing the experimental procedure, and Neerja Aggarwal and Amit Kohli for helpful feedback on this manuscript.

This project was funded by Packard Fellowship and Chan Zuckerberg Biohub Investigator Awards to Laura Waller and Daniel Fletcher, STROBE: An NSF Science \& Technology Center under Grant No. DMR 1548924, a NIH R01 grant to Daniel Fletcher, a NSF Graduate Research Fellowship awarded to Henry Pinkard, and a Berkeley Institute for Data Science/UCSF Bakar Computational Health Sciences Institute Fellowship awarded to Henry Pinkard with support from the Koret Foundation, the Gordon and Betty Moore Foundation through Grant GBMF3834 and the Alfred P. Sloan Foundation through Grant 2013-10-27 to the University of California, Berkeley.

\bibliography{references}
\bibliographystyle{unsrt}

\begin{appendices}
  \renewcommand{\thesection}{S\arabic{section}}
  \renewcommand{\thefigure}{S\arabic{figure}} 
  \setcounter{figure}{0} 

\section{Methods}
\label{section_methods}

In this section we described how the data were generated and processed into its final form.

\subsection{Sample preparation and imaging}

\paragraph{Assembling imaging chambers}

\begin{figure*}[htbp]
\centering
\includegraphics[width=\linewidth]{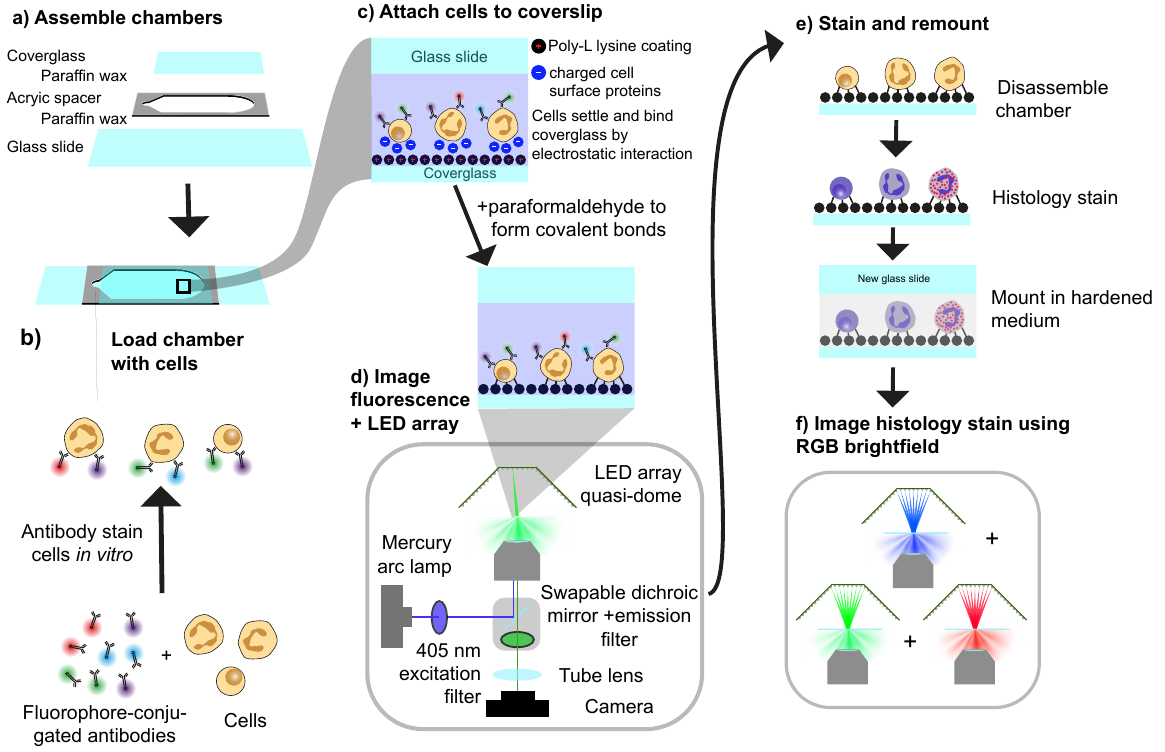}
\caption{\textbf{Sample preparation and imaging.} \textbf{a)} Imaging chambers were assembled by attaching an acrylic spacer between a microscope slide and cover glass using paraffin wax. \textbf{b)} Cells were stained with fluorophore-conjugated antibodies and loaded into the chamber by an opening at its end. \textbf{c)} Cells were attached to the coverslip first by binding through electrostatic interactions and then covalently using paraformaldehyde. \textbf{d)} The slide was then imaged using LED array and fluorescence illumination. \textbf{e)} After imaging, the slide was disassembled, a Wright's (histology) stain was applied, and the cells were mounted on a new slide with hardening mounting medium. \textbf{f)} RGB histology images were collected by illuminating with each color of the LED array in series.}
\label{experiment_procedure}
\end{figure*}

Imaging chambers were specially designed to 1) hold cells in an aqueous environment 2) be stored for a period of weeks in between cell isolation and imaging 3) be disassembled so that a subsequent histology staining step could be applied, without moving cells. Chambers were constructed with 600 $\mu$m acrylic spacer in between a poly-L lysine-coated \#1.5 coverslip and a standard glass microscope slide, creating a chamber with a volume of $\approx450 \mu$L (Fig. \ref{experiment_procedure}a). The shape of the spacer was empirically optimized for loading cells at one end of the chamber without the formation of unwanted bubbles, and it was cut from a larger sheet of acrylic using a laser cutter. The coverslips were cleaned with milliQ water then isopropanol then milliQ again, coated with poly-L lysine by placing them in a plasma cleaner, and then allowing 1mL of poly-L lysine solution to sit on the cleaned surface for one hour. The chambers were assembled by melting paraffin wax in onto the spacer and sandwiching it between two pieces of glass.

\paragraph{Cell isolation and staining}
12 mL of blood were drawn by venipuncture and added to 50mL tubes containing red blood cell lysis buffer (Fisher Scientific, \#NC9067514), which had been diluted with 10$\times$ deionized water as specified in manufacturer instructions. Cells were incubated for 10-15min at room temperature, while wrapped in aluminum foil to protect cells from light. Tubes were then centrifuged at 350g for 5 minutes. The supernatants were aspirated without disturbing the pellets, and all cells were then concentrated into a single 15 mL tube. This tube was filled with additional red blood cell lysis buffer and incubated at room temperature for another 10 min to get rid of remaining red blood cells. This tube was then centerfuged at 350g for 5 minutes and resuspended in 1mL IgG normal mouse serum control (Invitrogen \#PI31880) and put on ice. Cells were counted on a hemocytometer and resuspended at $2\times10^7$ per mL in IgG buffer.

Antibodies were added and samples stained on ice for 30 min. An additional 700 $\mu$L were added to each tube to fill them, and they were centerfuged at 350g for 5 min. The supernatant was discarded, and they were filled with PBS+EDTA, and centerfuged again using the same setting. The supernatant was discarded, and they were resuspended in 200$\mu$L PBS+EDTA and counted on a hemocytometer. Cells were then resuspended with 50-300k cells (depending on the number remaining in each tube) in 450$\mu$L of PBS+EDTA.

The following antibodies were used at the following concentrations : 

\begin{itemize}
    \item Brilliant Violet 510 anti-human CD19 \#302241 (5\% dilution)
    \item Brilliant Violet 570 anti-human HLA-DR \#307637 (6\% dilution)
    \item Brilliant Violet 605 anti-human CD45 \#368523 (5\% dilution)
    \item Brilliant Violet 711 anti-human CD16 \#302043 (4\% dilution)
    \item Brilliant Violet 510 anti-human CD56 \#318339 (6\% dilution)
    \item Brilliant Violet 510 anti-human CD3 \#317331 (5\% dilution)
    \item Brilliant Violet 421 anti-human CD123 \#306017 (5\% dilution)
    \item Brilliant Violet 650 anti-human CD14 \#301835 (5\% dilution)
\end{itemize}

\paragraph{Filling imaging chambers with cells}
Imaging chambers were filled by pipetting in suspended cells (Fig. \ref{experiment_procedure}b), and placed with coverglass facing down to allow the cells to settle on the coverglass and bind by electrostatic interaction. After 30 min, 22.5 $\mu$L were removed from the chamber using a pipette, taking care to keep the slide tilted and not let the resultant air bubble move throughout the chamber, as its surface tension tended to rip cells off the coverglass. Next, it was replaced with 22.5$\mu$L of 16\% paraformaldehyde to form covalent bonds between the cells and the coverslip. This process was repeated 4 times to increase the effective concentration of paraformaldehyde (Fig. \ref{experiment_procedure}c). The openings in the chambers were sealed to prevent evaporation by using epoxy to attach a small piece of acrylic to block the opening. Finally, the chambers were stored vertically (to prevent shear forces of air bubbles ripping away cells from the coverglass) at room temperature and protected from light until they were sequentially imaged over the next 2 weeks (Fig. \ref{experiment_procedure}d).

\paragraph{Histology staining}
After imaging in fluorescence and LED-array modes, some chambers were disassembled for histology staining (Fig. \ref{experiment_procedure}e). It was necessary to do the histology staining in a separate step, since having a histology stain on the cells during imaging with the LED array would alter their absorption/scattering properties, and could potentially alter the fluorescence signal as well. Imaging chambers were submerged under water while opened, in order to avoid the surface tension from air bubbles ripping cells off of the coverslip. The coverslip was first released from the acryclic space by sliding a razor blade underneath it. The top portion of it where the epoxied sealant was present was then removed by cutting off a piece of the coverslip with a diamond tipped knife. The slides were then stained with Wright's stain (Sigma-Aldrich \# 45253) by dipping in coplin jars (Grainger \#F44208-1000). They were submerged in the stain for one minute, followed by 5 dips in and out of a 75\% water 25\% ethanol phosphate buffer solution at pH 6.65. They were then dipped twice in a 75\% water 25\% ethanol low concentration (0.83 mM) phosphate buffer solution at pH 6.65. The 25\% ethanol was used to reduce the surface tension of the solution and minimize the chance of cells be dislodged from the coverslip.

The remaining stain on the slide was blotted off using a Kimwipe, and the slides were left to dry overnight. The next day, excess wax was scraped off the coverslip with a razor blade, 2 drops of anhydrous mounting medium were added (Milliport Neo-Mount \#109016), and a fresh microscope slide was attached. After drying, the top surface of the coverslip was cleaned with methanol prior to imaging (Fig. \ref{experiment_procedure}f). 

\paragraph{Microscope and data collection}

Samples were imaged on a Zeiss Axio Observer microscope using its standard fluorescence illumination path and its trans-illumination lamp replaced by a programmable quasi-dome LED array~\cite{Phillips2016}. The fluorescence excitation source was a mercury arc lamp with a single band-pass filter selecting for the 405 nm peak. Fluorescence and LED array images were collected with a 20$\times$ 0.5 NA air objective and histology images were collected using a 63$\times$ 1.4 NA oil objective. All images were taken on a Basler Ace acA2440-75um USB3 Monochrome camera, using the central 2056$\times$2056 pixel region. 

The microscope was controlled using Micro-Magellan\cite{Pinkard2016} with some additional customized modifications to its source code. These modifications have since been generalized, and are now part of Pycro-Manager \cite{Pinkard2021}. Micro-Magellan's explore feature was used to map out the imaging chamber, and its surface feature was used to mark the approximate position of the cells. 

Full scanning a single slide took $\sim$16 hours, in part because many channels were collected, and also because there was a several second delay at each XY position in order to allow the XY stage to fully settle and prevent loss of resolution due to motion blur. Because of this long acquisition time, the focus could drift by tens of $\mu$m over the course of an imaging session away from it's originally marked position. The microscope was not equipped with a laser-based hardware autofocus system, so there was a need for an autofocus mechanism that could be executed quickly and many times over the course of imaging.  We developed a single-shot autofocusing method to accomplish this, which is described in full in chapter 2 and elsewhere \cite{Pinkard2019}. This autofocusing routine was executed at each XY position to provide a more precise focus after each move of the XY stage.

Slide scanning of the histology slides was performed in a similar fashion. However, the single-shot autofocus algorithm did not give satisfactory results on the histology sample. This was in part due to the much smaller depth-of-focus of the histology objective, but also may have been inherent to the algorithm itself. As a result, an alternative autofocus algorithm was developed, in which focal stacks were taken and the sharpest plane was computed from them. Since this was more time consuming, it was only performed on a subset of XY positions, and a running average was kept to interpolate the focus offset at other positions.

Histology stained samples are typically imaged with brightfield illumination using an RGB camera. In our case, we achieved the same effect by collecting the three channels independently with a monochrome sensor, by lighting up red, green, and blue brightfield LEDs sequentially.

\paragraph{Exposures}
\label{exposure_bleedthrough_comp_sec}
The use of multiple fluorophores that were all excited by a common wavelength meant that there was in many cases substantial bleedthrough from one fluorescent channel to another (we note that this was a cost-saving decision meant to minimize the number of excitation filters needed). In order to maximize our ability to later unmix data into measurements of the component fluorophores, we tuned the exposure that each fluorescence channel was collected with. First, examples of strongly stained cells were collected for each fluorophore, with a constant exposure over all channels. The summed fluorescence intensity of each cell yielded a 6-element vector, containing the brightness of each fluorophore in each channel. These vectors were each multiplied by a unique scalar, and then stacked together to form a 6$\times$6 matrix. Using the condition number of this matrix (ratio of largest to smallest eigenvalue) as an objective function, these multipliers were optimized by gradient descent to find the best multipliers, which were then used to determine the ratio between exposures of different channels.

Exposures for the LED array channels (including the histology imaging) were picked for each channel by empirically finding a setting that effectively made use of the camera's dynamic range.

\subsection{Raw data to single-cell crops}

\begin{figure*}[htbp]
\centering
\includegraphics[width=0.7\linewidth]{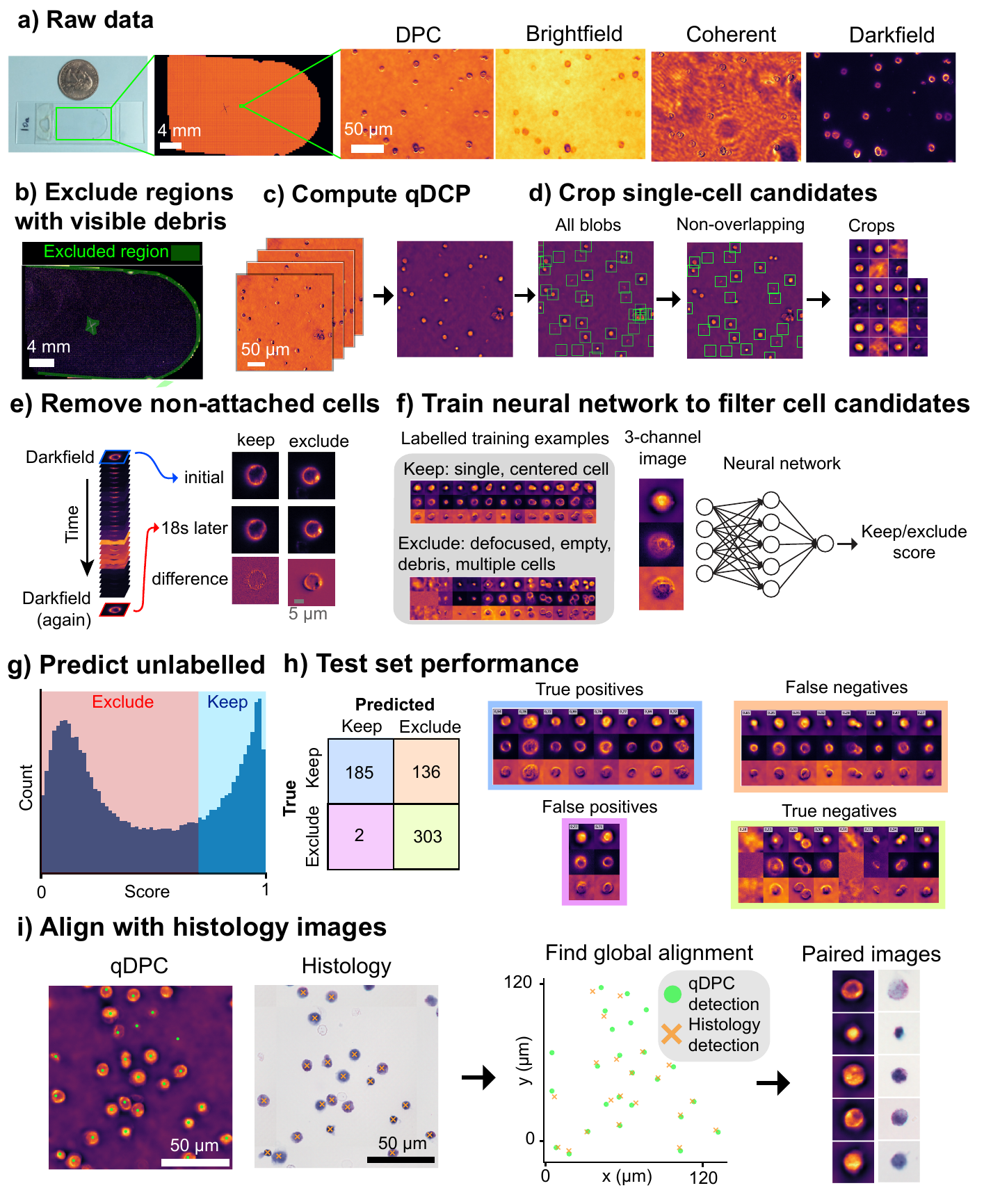}
\caption{\textbf{From raw data to single-cell crops} \textbf{a)} a single imaging chamber, an image a full slide scan, and zoom-ins in four different illumination patterns. \textbf{b)} Regions with visible debris were manually excluded from further processing. \textbf{c)} Quantitative differential phase contrast (qDPC) images were calculated for each field of view, and \textbf{d)} a blob-finding algorithm was employed to find and crop out candidate images for isolated single-cells. \textbf{e)} Candidates that were not attached to the coverslip, as measured by movement between the first and last darkfield image were removed. \textbf{f)} A manually labelled training set of cells to include or exclude was created and used to train a neural network that predicted whether to keep cells. \textbf{g)} Histogram of predictions on unlabelled cell candidates \textbf{h)} Performance of the trained network on the labelled test set. \textbf{i)} Detected cells in differential phase contrast and histology stain contrast were aligned and matched.  }
\label{data_processing}
\end{figure*}

A combination of manual data cleaning and machine learning was used to process the raw data from large slide-scans to a collection of single-cell crops. The raw data, full slide-scans in multiple channels (Fig. \ref{data_processing}a), were visualized using a custom-programmed multi-resolution viewer written in Python. This enabled the identification and exclusion of visible debris on the slide and areas near the edges where optical artifacts were visible (Fig. \ref{data_processing}b). Next, the 4 differential phase contrast (DPC) images at each field of view were used as input to an inverse algorithm that computed quantitative phase~\cite{Tian2015c} (qDPC) (Fig. \ref{data_processing}c). A difference of Gaussians blob-finding algorithm~\cite{Low2004} implemented in Scikit-Image~\cite{VanDerWalt2014} was used to identify candidates for single cells. The parameters of this algorithm were intentionally set somewhat permissively to be sure to capture small cells, and as a result many false positive were present and these cell candidates required extensive algorithmic filtering.

Since the end goal was to find isolated single cells, this filtering began by removing all cells that were too close to another detected cell (Fig. \ref{data_processing}d). In spite of the paraformaldehyde treatment, there were some cells that remained unsecured to the coverslip. This was clearly visible when watching a timelapse of the cells, as well as the fact that movement could be seen between different channels. To facilitate removal of these cells, the first and last channel during imaging ($\sim$18s apart) used the same darkfield illumination pattern on the LED array. By aligning these two images using a cross-correlation algorithm, cells that were moving could be removed (Fig. \ref{data_processing}e). Next, cells that were outside the fluorescence illumination fieldstop were excluded, so that all cells in the final dataset would have valid fluorescence measurements.

At this point, the cell candidates consisted of a mix of centered, in-focus single cells and empty areas, small acellular debris, out-of-focus cells, and clumps of cells (Fig. \ref{data_processing}f). A training set of $\sim$1000 cells was manually given binary labels for keeping and excluding. This labelled training set was then used to train a convolutional neural network capable of predicting whether unlabelled examples should be kept or discarded. The network architecture consisted of a 3 channel image (consisting of quantitative differential phase contrast image, a 0.5-0.6 NA darkfield image, and a brightfield image) fed into a DenseNet201~\cite{Huang2017} architecture, followed by a 400-unit fully connected layer with ReLu activation, a 0.5 probability dropout layer, another 400-unit fully connected layer with ReLu activation, another 0.5 probability dropout layer, and 2-unit fully connected layer with a softmax output. The network was trained in Keras\cite{chollet2015keras} with the Adam optimizer\cite{Kingma2015a} with a learning rate of $3\times10^{-6}$ and a batch size of 8. Training was continued until loss stopped decreasing on a held-out validation set. The performance of this network was optimized by using Keras-Tuner\cite{omalley2019kerastuner} to perform Bayesian optimization over its hyperparameters to achieve optimal performance on a held out set of validation data. 

Examining the performance of this network on a test set of held-out labelled examples, it was clear that many failure cases occurred when the network misidentified a clump of two cells as one. To compensate for this, the same blob finding algorithm used to originally locate cells, but with different parameters that erred on the side of detecting multiple cells when only one was present, was used to identify unlabelled cell candidate crops that potentially contained more than one cell. These examples were then labelled, thereby filling the training set with many more examples of multi-cell images, which improved its performance on these types of images on the test set. One factor that made this especially difficult was that many lymphocytes, during the time in between when they were isolated and fixed by paraformaldehyde, had begin to undergo apoptosis. The resultant expulsion of their cytoplasm gave them a characteristic dumbbell-like shape, which looked very much like two cells stuck together and was in some cases even difficult for the human-labeller. Because of this, the labels erred on the side of excluding these cases, and as a result cells that looked like this were more likely to be excluded from the final dataset. 

Finally, the trained network was applied to all unlabelled data to score each cell candidate for whether it should be removed or kept. Labelling cell candidates for keeping or removing was ultimately subjective, especially when it came to how well in focus each cell was. Because of this, after labelling a certain amount of training data, classification accuracy on the test set showed asymptotic improvement with more data at around $\sim$80\% accuracy. This behavior suggested that no further improvement was possible given the inherent noise in the human-provided labels. In addition, most errors on the test set appeared to be drawn from the harder to distinguish examples where this noise would be expected to be most prominent. We decided it was more important to reduce false positives than false negatives in the final dataset. That is, excluding non-centered/out of focus/non-cell images was more important than ensuring all high-quality images included. Thus, we used a conservative cutoff for the keep/exclude score of 0.7. Figure \ref{data_processing}g shows the histogram of predictions on unlabelled data, as well as results on the test set.

Cells in the histology images were identified by a similar procedure. Cell candidates were first identified by a difference of Gaussians blob finding algorithm. Next, these candidates were filtered by summing all the gradients in each image and removing the ones low-values, which reliably removed out-of-focus cells or empty crops. This was the only exclusion step needed, since finding a global alignment over the full slide between histology and LED array/fluorescence images enabled one-to-one matching of histology cell candidates to LED array/fluorescence ones, and the latter had already been filtered to match the desired criteria. The histology and LED-array/fluorescence slides were aligned by computing an objective function that measured how close each histology cell was to a corresponding fluorescence cell, given global translation and rotation parameters applied to all detected cells. Specifically, for each histology cell the closest LED array/fluorescence cell was found, and the sum of these minimum distances over all histology cells was computed. A grid search over all possible values of translation and rotation of the two slides was performed, using Jax\cite{jax2018github} for GPU-acceleration. The optimal value was confirmed by looking at the visual alignment of the two contrast modalities and the structurally similar features between the two contrast modalities (Fig. \ref{data_processing}h).

\subsection{Fluorescence processing/demixing}
\label{demixing}

\begin{figure*}[htbp]
\centering
\includegraphics[width=\linewidth]{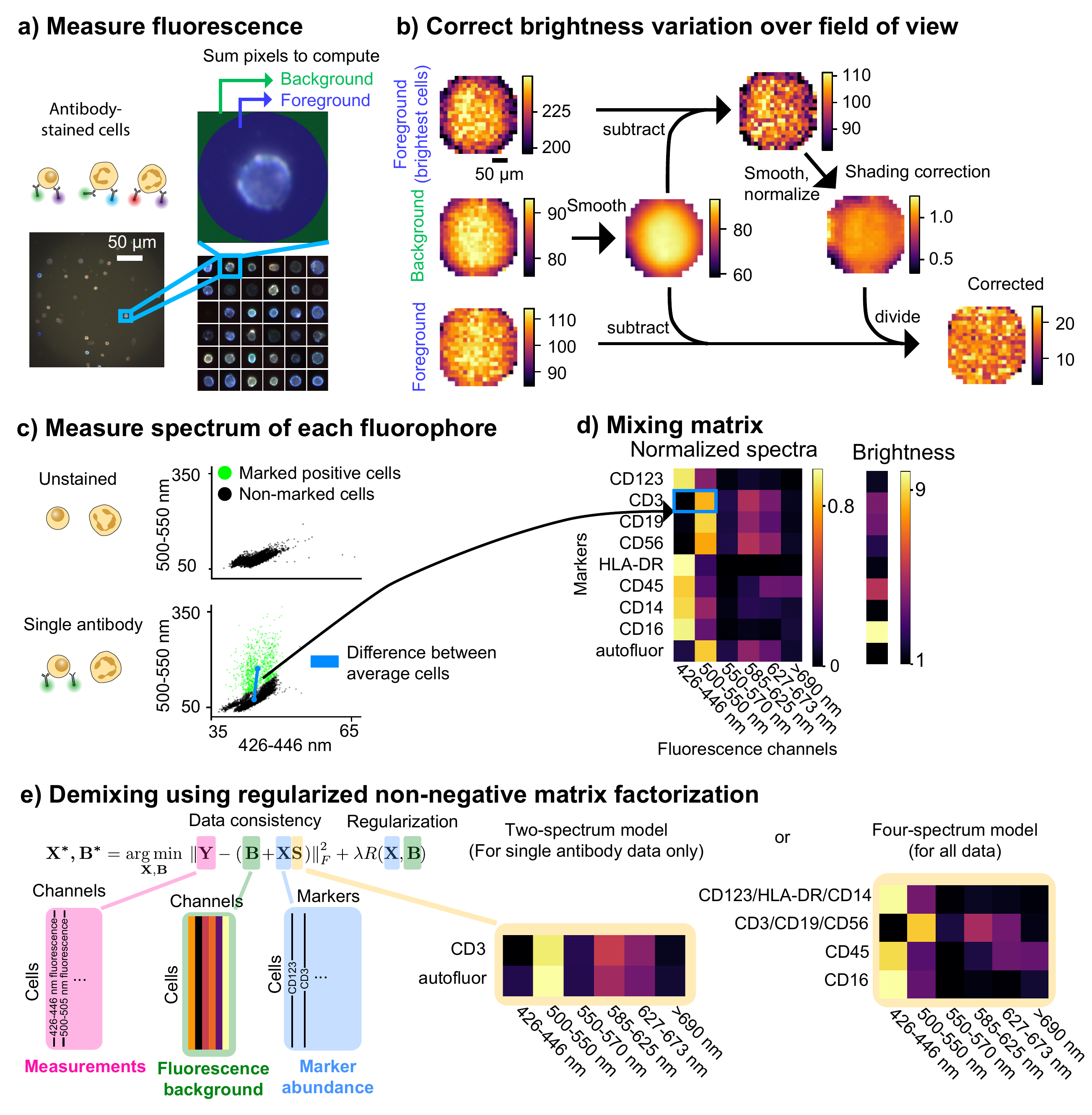}
\caption{\textbf{Raw fluorescence to protein estimates} \textbf{a)} Fluorescent cells in a single field of view, and the areas of each crop used to compute foreground and background fluorescence estimates. \textbf{b)} The background subtraction and shading correction procedure used to correct for spatial variation in brightness across the field of view. \textbf{c)} The spectrum of each fluorophore was computed by looking at cells stained with the corresponding antibody vs. no antibodies and taking difference of the means of antibody-positive and antibody-negative cells. \textbf{(d)} The normalized spectra and relative brightness for each antibody and the autofluorescence. \textbf{e)} The regularized non-negative matrix factorization optimization problem that was solved to give estimated of the relative abundance of each protein. This problem utilized a two-spectra model (for single antibody conditions) or a four-spectra model (for every antibody condition)}
\label{fluor_and_demix}
\end{figure*}

After computing a finalized set of single, isolated, in-focus cells, the fluorescence images of each of these cells was processed in order compute estimates of the relative amounts of the protein targets of each antibody. First the raw fluorescence was measured in the foreground and background of each cell by summing all the pixels inside or outside (Fig. \ref{fluor_and_demix}a) of a circle inscribed in the square crop. This size was chosen so that the pixels outside of it captured the local background intensity of that area of the slide/field of view, while the pixels inside contained the background plus fluorescence from the cell. Background subtraction and shading correction are a common practice in quantitative fluorescence microscopy due to the nonuniform pattern of excitation light across the field of view as well as the nonuniform collection of fluorescence on the edges of the image\cite{Smith2015, Peng2017}. In these data, there were also variations in background fluorescence between slides, likely due to antibodies in the chamber that were unbound to cells. The spatial pattern of the background for each slide was first visualized using a spatial histogram, and then a smoothed version of this was calculated using a locally weighted scatterplot smoothing (LOWESS) using locally linear regression \cite{Cleveland1979} (Fig. \ref{fluor_and_demix}b). The foreground brightness was calculated using the same procedure on the brightest 10\% of cells after subtracting the smoothed background, yielding a shading correction. Taking the brightest cells gave a better estimate since only a fraction of cells actually had antibody for a given fluorophore bound. Finally the foreground of all cells was corrected for spatial variations by subtracting the smoother background and dividing by the shading correction, yielding a spatial histogram that no longer displayed obvious spatial variation (Fig. \ref{fluor_and_demix}b, bottom right). This process was repeated over each fluorescent channel and both batches of data.

These spatially corrected fluorescence measurements were then used to solve a demixing inverse problem--going from fluorescence intensities to the relative levels of antibody-bound proteins or autofluorescent molecules that gave rise them. To solve this problem the spectra of the various fluorescent species were measured by comparing scatter plots of cells' fluorescence in conditions where they were treated with a single antibody vs the condition were they were treated with no antibody (Fig. \ref{fluor_and_demix}c). Taking the difference of the means of the antibody-positive and antibody-negative populations gave a vector that measured that fluorophores' spectrum. Cells also showed varying amounts of autofluorescence. To capture the spectrum of this autofluorescence, the same procedure was repeated with the brightest autofluorescence and the rest of the population with these excluded. These measured spectra were then used to form "mixing matrices", which could be used as part of the demixing problem (Fig. \ref{fluor_and_demix}d).

\paragraph*{Fluorescence unmixing with non-negative matrix factorization}

Computing a spectral unmixing inverse problem, in which the levels of individual proteins can be found from the levels of overlapping spectra, can be posed as a non-negative matrix factorization problem. This models the physics of the imaging process since neither the spectra nor the fluorphore density can be negative. A simplified version of the optimization problem can be seen in Fig.~\ref{fluor_and_demix}e. The exact problem including normalizing constants was:

$$
\boldsymbol{\mathrm{X}^*, \mathrm{B}^*} = 
\underset{\mathbf{X}, \mathbf{B}}{\arg\min} \: 
\frac{1}{NC}\left\lVert \mathbf{Y} - (\mathbf{X} \mathbf{S} + \mathbf{B}) \right\rVert _F^2 +
\lambda \frac{1}{NM} \left\lVert \text{vec}(\mathbf{X} \textrm{diag}(\mathbf{w}))  \right\rVert _1 +
\beta \frac{1}{NC} \left\lVert \text{vec}(\mathbf{B})  \right\rVert _1
$$

Where 
$$
\left\lVert\boldsymbol{\mathrm{A}}\right\rVert_{F}^2 = \sum_{i,j} a_{ij}^2 \ \ \ \  \textrm{(Frobenius norm)}
$$

$$
\left\lVert\text{vec}(\boldsymbol{\mathrm{A}})\right\rVert_1 = \sum_{i,j} |a_{ij}| \ \ \ \ \text{(Elementwise L1 norm)}
$$


\noindent $N$ is the number of cells, $C$ is the number of fluorescence channels, and $M$ is the number of spectra (i.e. one for each unique fluorophore or group of similar fluorophores). $\mathbf{X}$ is an $N \times M$ matrix where each row contains the levels of each protein for a given cell, $\mathbf{Y}$ is an $N \times C$ matrix containing the observed fluorescence of each protein for a given cell, $\mathbf{S}$ is an $M \times C$ matrix containing the fluorescence spectrum for each protein as a row, and $\mathbf{B}$ is an $N \times C$ matrix containing the background level fluorescence in each channel. Each row of $\mathbf{B}$ was constrained to be identical for cells from the same physical microscope slide, thereby enforcing the constraint of a global level of background fluorescence specific to each slide. 

$\textrm{diag}(\mathbf{w})$ is a diagonal matrix formed by putting the entries of the $M \times 1$ column vector $\mathbf{w}$ along the diagonal, where $\mathbf{w}$ is a weighting vector that enables regularizing the levels of different fluorophores independently. This was useful since different fluorophores varied by over an order of magnitude of absolute brightness. We found that a useful heuristic for setting the value of each element of $\mathbf{w}$ was to project the normalized spectrum of each fluorophore onto the first right singular vector of $\mathbf{S}$ and divide the result by spectrum's magnitude.

$\lambda$ and $\beta$ are global regularization tuning parameters for $\mathbf{X}$ and $\mathbf{B}$, respectively, and have values: $\lambda=7\times10^{-1}$ $\lambda=5\times10^{-2}$.

The optimization problem was solved using gradient descent with momentum with a learning rate of $1\times10^3$ and a momentum of 0.9. This was implemented computationally using Jax~\cite{jax2018github}.

An important choice is which spectra will be included in the unmixing matrix. For the cases in which only a single antibody was used, more accurate results can be obtained by building this knowledge into the optimization problem and excluding spectra of fluorophores that aren't present, leaving only the antibodies spectrum and the ever-present autofluoescence spectrum forming a two-spectra mixing matrix (Fig. \ref{fluor_and_demix}e). 

In the case where the cells were stained with all antibodies at once, this problem becomes more complex. Ideally all of the spectra will be linearly independent and there would be more measurements (fluorescent channels) than fluorescent species to unmix. For these data, neither of these are true. Some fluorophores are similar in spectra, and the presence of autofluorescence in addition to the six fluorophores means that this is an underdetermined problem, with only six measurement channels. To compensate and come to a reasonable solution, it is necessary to combine multiple similar spectra with insufficient signal-to-noise to be individually distinguished into one. The spectra for CD14, HLA-DR, and CD123 were combined into a single spectrum in order to achieve this. This is only possible because of the similarity among these spectra (Fig. \ref{fluor_and_demix}d). Though the CD16 fluorophore also appears to have a similar spectra to these three, we note that its faint fluorescence in the $>$690 nm channel, combined with it having the highest absolute brightness enabled it to be effectively identified by the algorithm. CD3, CD19 and CD56 were also combined into a single channel, since the antibodies that targeted them all used the same fluorophore. Merging spectra in this way yielded the four-spectrum mixing matrix (Fig. \ref{fluor_and_demix}e). 

\begin{figure*}[htbp]
\centering

\includegraphics[width=0.9\linewidth]{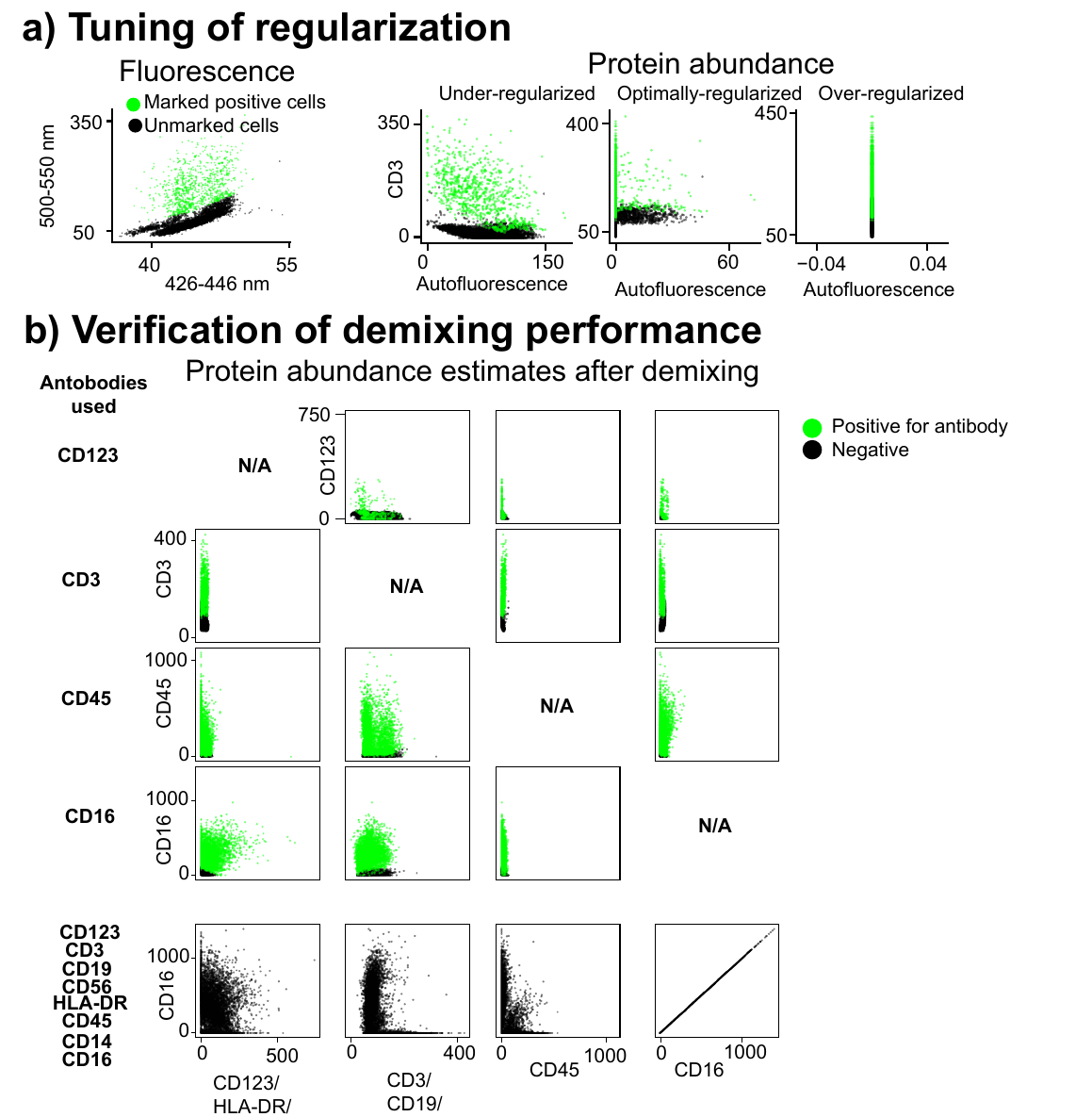}
\caption{\textbf{Analysis of demixing performance} \textbf{a)} The effect of choosing different regularization levels on the two spectrum model. Under-regularizing fails to separate marked antibody-positive cells (green) from unmarked cells (black). Over-regularizing separates the two, but collapses all autofluorescence values to 0. Optimally regularizing balances these two. \textbf{b)} The 4-spectrum demixing model applied to single-antibody stained data (top 4 rows) or all antibody stained data (bottom row). For the single-stain cases, the algorithm successfully separates marked cells from non-marked cells with only small estimated amounts for antibodies not present in most cases, though there is there is some error for certain antibodies: for example, CD16 and CD45 into the CD3/CD19/CD56 channel}
\label{demix_detailed}
\end{figure*}

For the two-spectra model, validating the correct regularization level was performed by visualizing how well a known antibody-positive population (Fig. \ref{fluor_and_demix}a) is separated from a known antibody-negative population (Fig. \ref{demix_detailed}a). Ideally, a horizontal line separating these populations could be drawn (Fig. \ref{demix_detailed}a, optimally-regularized). If the regularization is too weak, this line no longer runs perpendicular to the axis of the unmixed protein (Fig. \ref{demix_detailed}a, under-regularized). If too strong, the levels of the second fluorescent species are all zero  (Fig. \ref{demix_detailed}a, over-regularized).

For the four-spectra unmixing model, results were validated by unmixing data that came from a single antibody staining condition. In this scenario, it is known that the correct answer should assign variation only along the the axis of the protein target of that particular antibody. Figure \ref{demix_detailed}b shows the result of this experiment for four single-antibody conditions corresponding to the four spectra in the top four rows, as well as the condition with all antibodies in the bottom rows. This experiment shows that some unmixing results using this model were very reliable. For example, this can be seen in the fact that unmixed CD3 only data barely registered non-CD3 signal, and similarly for the cross-talk between CD45 and CD16. CD123, CD45 and CD16 all gave rise to signal in the CD3/CD19/CD56 spectra, which may be due in part due to the face that this model didn't explicitly account for autofluorescence.

After developing and validating these models, the unmixing inverse problem was solved on all applicable data to obtain the levels of protein abundance on each cell. The two-spectrum model was applied to all cells stained with a single antibody (with the spectrum of that antibody inserted as appropriate). The four-spectrum model was applied to all data. Though two-spectrum results are likely more accurate when available, single antibody staining conditions were also unmixed with the four-spectrum for comparison purposes.

\subsection{Defining cell types for classification}

\begin{figure*}[htbp]
\centering
\includegraphics[width=1\linewidth]{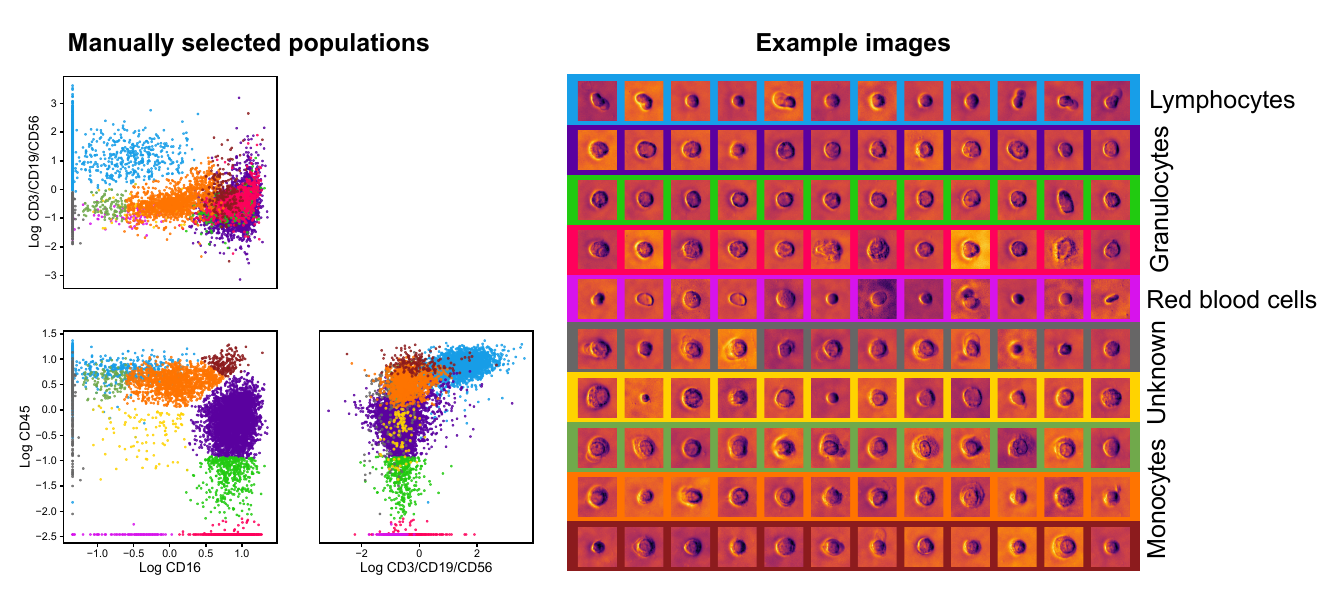}
\caption{\textbf{Protein expression levels, example images, and putative cell types of the classification labels}  }
\label{classification}
\end{figure*}

Labels for cell type classification were generated by manually selection subpopulations of the all antibody stained cells in batch 0. Cells were selected based on their expression of CD45, CD16, and CD3/CD19/CD56. The CD123/CD14/HLA-DR channel was deemed too noisy be useful. \textbf{Figure \ref{classification}} shows how cells were delineated into classes based on their expression level, example differential phase contrast images of the cells, and the putative cell types each class belongs to. The dataset contains two different versions of the labels, one with only Lymphocytes/Granulocytes/Monocytes, and one with 10 classes that include subtypes of these three along with unknown cell types and red blood cells. It is unclear if all of these 10 classes are related to biological variability, rather than noise in the measurement procedure.

\subsection{Known imperfections}

In the construction of this dataset, several technical errors were made along the way, which we describe here:

\begin{itemize}
    \item For the cells in batch 1 that were stained with all antibodies, the amount CD19 antibody used was 35 percent of the amount used in other CD19 stains
    \item There was a small drop of oil on an internal lens element in the objective lens used for the LED-array/fluorescence imaging. When imaging with certain low-NA single LEDs, this created a strongly visible artifact across the field of view. Because of this, these LEDs are excluded from BSCCM-coherent.
    \item All data were imaged with a region of interest (ROI) set on the camera to a central 2056$\times$2056 region. However, for cells in the no antibody, batch 1, replicate 1 slide, this ROI was not set. This may have caused differences in fluorescence exposure and spatial variations compared to other slides.
    \item As discussed in Section \ref{exposure_bleedthrough_comp_sec}, the fluorophores/filters chosen were not ideal for multi-channel imaging. This can be compensated somewhat as described in that section, but a far better choice would have been to use fluorophores with unique excitation spectra and their own matched excitation filters. We opted not to do this in an effort to minimize cost. We note that hindsight is 20/20.
    \item From slide-to-slide and batch-to-batch, there are variations in the average intensity of images in a given channel. It is not known whether this resulted from variations in the camera or the LED array. For example, the LED119 channel, though collected with the same settings on very similar cells, has very different average brightness between batch 0 and batch 1.
\end{itemize}

\section{Data organization}
\label{section_data_and_files}

This section describes how the data is organized: which files contain which data, what metadata is available, etc. We note that a full understanding of these details is not necessary for using the dataset, as the Python package we provide abstracts away many of these implementation details.

\subsection*{File structure and organization}

All image data are stored in Zarr \cite{https://doi.org/10.5281/zenodo.3773449} datasets using Blosc/zstd compression. Tabular metadata (i.e. per-cell metadata) are stored in .csv files. Global metadata, which contains information that is not specific to individual cells, but rather pertains to the whole dataset is stored in text files in Javascript Object Notation (JSON) format.

Each top-level BSCCM (regular, coherent, tiny, or coherent-tiny) contains (up to) 5 items:

\begin{itemize}
    \item \textbf{BSCCM\_images.zarr}: A Zarr dataset containing all the images of cells
    \item \textbf{BSCCM\_backgrounds.zarr}: A Zarr dataset containing the background intensity over the full field of view, for each LED array illumination pattern. 
    \item \textbf{BSCCM\_global\_metadata.json}: A text file containing metadata about the full dataset (pixel size, wavelength, channel names, etc.) in JSON format
    \item \textbf{BSCCM\_index.csv}: A comma separated value (CSV) file containing metadata specific to each cell in the dataset
    \item \textbf{BSCCM\_surface\_markers.csv}: A comma separated value (CSV) file containing information about the surface protein marker levels of each cell, along with many measurements derived from the fluorescence images and intermediate values used in computing these levels
\end{itemize}

\paragraph{BSCCM\_images.zarr}
Zarr datasets contain a hierarchy of directories. For the BSCCM\_images.zarr file, this has the following structure:

\begin{lstlisting}

+-- antibodies_CD16
|    +-- batch_0
|    |    +-- slide_replicate_0
|    |        +-- dpc
|    |            +-- cell_0
|    |            +-- cell_1
|    |            ...
|    |        +-- fluor
|    |            ...
|    |        +-- led_array
|    |            ...
|    |        +-- histology
|    ...
+-- antibodies_CD45
...
\end{lstlisting}

The outermost directory contains ``antibodies\_" followed by the name of the antibody used to stain the cells, or ``unstained"/``all" for the no-antibody and all-antibody conditions, respectively.

The next level contains directories named ``batch\_" followed by the batch index and is either 0 or 1 for BSCCM/$\text{BSCC\color{OliveGreen}{MNIST}}$ or 1 for BSCCM-coherent. The batch index captures the two biological replicates (i.e. distinct cell isolations) on two different dates. There is presumably some degree of biological variation between these two isolations, in addition to variation in antibody staining/processing/etc.

The next level contains directories named ``slide\_replicate" followed by a 0 or 1. Due to the number of available cells, some conditions (i.e. a given batch/antibody) were split among multiple physical microscope slides and imaged on different dates. The vast majority of cells/slides did not have a second replicate.

The next level contains directories that identify the contrast modality and is one of:
\begin{itemize}
    \item ``dpc": Quantitative differential phase contrast images (which were computed from raw DPC led-array images)
    \item ``fluor": Fluorescence images, either from the fluorphores attached to antibodies, or the cells' inherent autofluorescence
    \item ``led\_array": Images taken with different LED array illumination patterns
    \item ``histology" RGB images of histology stained cells
\end{itemize}

The next level contains ``cell\_" followed by the cell's global index. Each cell in each dataset has a unique global index, which allows them to be matched with per-cell metadata (described below).

Finally, within each cell directory are the Blosc/zstd-compressed binary data, split into individual files per each channel in order to maximize performance when reading only a single channel.

\paragraph{BSCCM\_backgrounds.zarr}

This file contains the background images for each channel across the full field of view (2056x2056 pixels). The top level directory contains the channel name. The subdirectories are  different versions depending on which pixel-wise percentile was taken over 200 images, with possible options of 5, 10, 20, 40, and 50 (median). The structure is as shown below.

\begin{lstlisting}
+-- Brightfield
|   +-- 5\_percentile
|   +-- 10\_percentile
|   +-- 20\_percentile
|   +-- 40\_percentile
|   +-- 50\_percentile
+-- DF\_50
...
\end{lstlisting}

\paragraph{BSCCM\_global\_metadata.json}

This file contains metadata specific to the full dataset, including names of channels, collection settings like exposure, and useful information for calibration like wavelength, objective NA, etc. It is a text file with JSON structure, as shown below:

\begin{lstlisting}
{
"led_array": 
    {
    "image_shape": [128, 128],
    "channel_names": ["Brightfield", ... "LED119"], 
    "channel_indices": {"Brightfield": 0, ... "LED119": 200}, 
    "exposure_ms": {"Brightfield": 8, ... "LED119": 200},
    "camera": {
        "offset": 30,
        "gain_db": 4,
        "quantum_efficiency": 0.68
        },
    "wavelength_nm": 515, 
    "pixel_size_um": 0.166, 
    "objective": {"NA": 0.5, "magnification": 20}}, 
"fluorescence": 
    {
    "image_shape": ...
    ...
\end{lstlisting}

\paragraph{BSCCM\_index.csv}

This contains per-cell metadata in a single, large CSV file with one row per each cell and the following columns:

\begin{lstlisting}

"global_index": An integer uniquely identifying the cell
"position_in_fov_y_pix"/"position_in_fov_x_pix":Location of the cell center within the image field of view
"detection_radius": The radius reported by the blob finding algorithm that initially located the cell, which gives a rough estimate of its size
"has_matched_histology_cell": Whether or not the cell has a matching cell in histology contrast
"fov_center_x"/"fov_center_y"/"fov_center_z": the microscope stage coordinates of the field of view from which the cell was drawn
"batch": the index of the cell isolation experiment the cells were drawn from (either 0 or 1)
"antibodies": the name or the single antibody used to stain the cells, or 'all' or 'unstained' if every antibody or no antibodies were used
"imaging_date": the date the slide of cells was imaged
"data_path": the path to the image data with the BSCCM\_images.zarr file
"slide_replicate": the index of the slide replicate within the same antibody/batch conditions (either 0 or 1)
\end{lstlisting}

\paragraph{BSCCM\_surface\_markers.csv}

This contains per-cell calculations about fluorescence surface marker levels. It is entirely derived from the fluorescence imaging data (as described in the methods). It contains metadata in a single, large CSV file with one row per each cell and the following columns:

\begin{lstlisting}
"global_index": An integer uniquely identifying the cell
// raw measurements derived from fluorescence images
"Fluor_426-446_total_raw" //Raw foreground fluorescence in 426-446nm channel
"Fluor_500-550_total_raw" //Raw background fluorescence in 500-550nm channel
...
"Fluor_426-446_background" //Raw background fluorescence in 426-446nm channel
...
(Many more intermediate calculations)
...
// protein levels estimates from unmixing procedure
"CD45_single_antibody_model_unmixed" // CD45 protein levels using 2 spectrum unmixing model
"CD123_single_antibody_model_unmixed"
...
"CD123/HLA-DR/CD14_full_model_unmixed" // Combined CD123/HLA-DR/CD14 protein levels using 4 spectrum unmixng model
...

\end{lstlisting}



\end{appendices}

\end{document}